%% file: lrec2022-example.tex
\documentclass[10pt, a4paper]{article}
\usepackage{lrec2022} 
\usepackage{multibib}
\newcites{languageresource}{Language Resources}
\usepackage{graphicx}
\usepackage{tabularx}
\usepackage{soul}

\usepackage{import}
\usepackage{titlesec}
\titleformat{\section}{\normalfont\large\bfseries\center}{\thesection.}{1em}{}
\titleformat{\subsection}{\normalfont\SmallTitleFont\bfseries\raggedright}{\thesubsection.}{1em}{}
\titleformat{\subsubsection}{\normalfont\normalsize\bfseries\raggedright}{\thesubsubsection.}{1em}{}
\renewcommand\thesection{\arabic{section}}
\renewcommand\thesubsection{\thesection.\arabic{subsection}}
\renewcommand\thesubsubsection{\thesubsection.\arabic{subsubsection}}

\usepackage{epstopdf}
\usepackage[utf8]{inputenc}

\usepackage{hyperref}
\usepackage{xstring}

\usepackage{color}

\title{Using Synthetic Data for Conversational Response Generation in Low-resource Settings}

\name{Gabriel Louis Tan, Adrian Paule Ty, Schuyler Ng, Denzel Adrian Co \\ {\bf \large Jan Christian Blaise Cruz \textnormal{and} Charibeth Cheng}}

\address{Center for Language Technologies (CeLT), De La Salle University \\ 2401 Taft Ave., Malate, Manila, Philippines \\ 
        \{gabriel\_louis\_tan, adrian\_ty, schuyler\_ng, denzel\_adrian\_co, jan\_christian\_cruz, charibeth.cheng\}@dlsu.edu.ph\\ }

\abstract{
Response generation is a task in natural language processing (NLP) where a model is trained to respond to human statements. Conversational response generators take this one step further with the ability to respond within the context of previous responses. While there are existing techniques for training such models, they all require an abundance of conversational data which are not always available for low-resource languages. In this research, we make three contributions. First, we released the first Filipino conversational dataset collected from a popular Philippine online forum, which we named the “PEx Conversations Dataset”. Second, we introduce a data augmentation (DA) methodology for Filipino data by employing a Tagalog RoBERTa model to increase the size of the existing corpora. Lastly, we published the first Filipino conversational response generator capable of generating responses related to the previous 3 responses. With the supplementary synthetic data, we were able to improve the performance of the response generator by up to 12.2\% in BERTScore, 10.7\% in perplexity, and 11.7\% in content word usage as compared to training with zero synthetic data.
 \\ \newline  }

\begin{document}

\maketitleabstract
\section{Introduction}

Existing language models with real life applications require huge amounts of data to produce favorable results or an improvement in performance. A customer service chatbot trained a Seq2Seq model with long short-term memory networks on over 1M Twitter brand accounts \cite{xu2017new}; a commercial artificial intelligence by Amazon, Alexa, was  pretrained on 4M utterances with each domain fine-tuned with 50K grammar utterances \cite{goyal-etal-2018-fast}. The assumption of these extensive resources poses challenges for low-resource languages with little to no existing dialogue corpora.

As opposed to attempting to construct extensive conversational datasets, data augmentation (DA) techniques can be used to alleviate the deficiency by generating synthetic data using pre-existing data. In NLP, one prevailing technique is word replacement where a selected word can be replaced with synonymous words \cite{Wei-etal-2019-EDA}, morphologically similar words \cite{vania-etal-2019-systematic}, or entities of the same type \cite{Dai2020}. Another technique is modifying instances by using correct sentences and applying a set of transformations that introduce errors for grammar correction \cite{grundkiewicz-etal-2019-neural}. 

In response generation, the work of \newcite{puri-2020-TrainingQA} shows that question answering systems can be successfully trained entirely by using synthetic data using an approach that involves generating questions and answers from a GPT-2 model. \newcite{baheti-etal-2020-fluent} explored fluent response generation to conversational questions which utilized Syntactic Transformations on the SQuAD 2.0 QA dataset to extract supervised conversational training data for response generation models based on Pointer-Generator Networks (PGN) \cite{see-etal-2017-get} and DialoGPT \cite{zhang2019dialogpt}. 

These previous works show that models perform better with large datasets, and even when they lack the resources, synthetic data allows them to perform just as well. However, these assume the existence of these resources in the targeted language, whether it be proven methods or datasets. All works mentioned using data augmentation primarily used English sources. This makes it difficult to adapt these techniques into low-resource languages like Filipino.

In this work, we constructed the first Filipino conversational dataset from an online Filipino forum website called Pinoy Exchange (PEx). With this dataset, we generated synthetic data through RoBERTa-based data augmentation, experimenting with the quantity of token replacements in an utterance. Then with varying experimental setups involving the token replacement percentage and simulated low data sizes, we fine-tune a DialoGPT model for a conversational response generator. We benchmark our models using perplexity, content word usage, and BERTScore, an automatic evaluation metric that favors contextual embedding matching over string matching.

\section{Filipino Conversational Dataset}

    
    


    

In order to alleviate the absence of conversational data in Filipino, we produce the PinoyExchange Conversations Dataset or "PEx Conversations".
The dataset consists of 2,424,748 comments across 45,621 threads, producing a total of 2,171,579 conversation-like exchanges.
    
Conversational data was extracted from PinoyExchange \footnote{\url{https://www.pinoyexchange.com/}} (PEx), an online forum mainly used by Filipinos with topics concerning the Philippines or Filipinos in general. The dataset is primarily in casual Filipino, with the addition of some English words commonly used in daily Filipino conversations. 

We construct the dataset by scraping the Small Talk subforum and the different subforums within the lifestyle category namely Food and Drinks, Home and Garden, Style and Fashion, Travel and Leisure, Visas and Immigration, Health and Wellness, and Body and Fitness. All unicode errors are resolved. Media-related information, links, account tags, emojis, repetitive punctuations, and thread-related nuances (i.e. referencing) were removed. Following \newcite{cruz2021exploiting}, Further cleaning methods such as decapitalization, removal of stopwords, stemming and lemmatization, and spelling and grammar correction were not performed to preserve the conversational nuances. 


\section{Methods}
\subsection{RoBERTa-based Data Augmentation}

Given that other existing techniques require external resources such as WordNet, \newcite{kobayashi-2018-contextual} introduces contextual augmentation, a technique which uses bi-directional language models to consider context in order to better predict words for word replacement tasks. Due to its memory capabilities for long term dependencies, BERT has been explored as a viable option for data augmentation as it is frankly more powerful than ordinary bidirectional language models \cite{wu-etal-2018-bert}.

Bidirectional Encoder Representations from Transformers (BERT) 
\footnote{\url{https://github.com/google-research/bert}} is a multi-layer bidirectional Transformer encoder that is designed to pretrain deep bidirectional representations that can be fine-tuned to a wide range of tasks \cite{devlin-etal-2019-bert}. BERT is pretrained on the masked language modelling (MLM) task and the next sentence prediction (NSP) task. The model’s capability to contextually-fill masked tokens opens an opportunity for the creation of synthetic data.

\textbf{RoBERTa.} \newcite{liu2019roberta} found that the original BERT pretraining procedure  was significantly undertrained and proposed a tweaked approach of training BERT models that improve end-task performance, which they referred to as Robustly optimized BERT approach (RoBERTa). RoBERTa was trained longer with dynamic masking, sentences without NSP loss, larger batches, and a larger byte-level Byte-Pair Encoding (BPE) and achieved state-of-the-art results on the GLUE, RACE, and SQuAD datasets. 

\subsection{DialoGPT}

DialoGPT is a response generation model based on the GPT-2 transformer architecture which uses masked self-attention and feedforward neural networks to predict the probability of the next output token. DialoGPT is trained using a 147M multi-turn dialogue dataset scraped from Reddit discussion threads,  leveraging a stack of masked multi-head self-attention layers in generating realistic responses. Specifically, DialoGPT inherits the transformer with layer normalization, the same initialization scheme but modified, and the same BPE for the tokenizer. Although the model is initially trained on an English corpora, DialoGPT is an open-domain pretrained model that can be further fine-tuned on a language-specific dataset to obtain a custom conversational model in any language such as Spanish \footnote{\url{https://huggingface.co/ncoop57/DiGPTame-medium}}.

\subsection{Evaluation Metrics}

Perplexity is a simple but powerful metric, a standard in most NLP tasks. It is the inverse probability of the test set, normalized by the number of words. It analyzes the probability of n-grams (product across all consecutive n-grams) appearing as the model learned in the dataset. Perplexity is normalized so that it does not rely on the size of the dataset, to give a more accurate per-word measure. Perplexity has been shown to correlate very well with how humans perceive coherent and natural conversations \cite{adiwardana2020humanlike}.

BERTScore is a text generation evaluation metric introduced by \cite{zhang2020bertscore}. The automatic evaluation metric computes for the similarity score for each token matched in the reference set, similar to common n-gram matching metrics such as BLEU, METEOR, and NIST. However, it has been proven that these common metrics are restrictive with regards to matching words with multiple meanings due to the popular approach of string-level matching. BERTScore addresses the issue by computing the similarity score based on BERT's contextual embeddings. Contextual embeddings are able to preserve the meanings of similarly-spelled words in varied contexts, and have been shown to work for tasks such as paraphrase detection \cite{devlin-etal-2019-bert}. The sum of cosine similarities between token embeddings is computed for the similarity score between two sequences. Additionally, BERTScore has also been found to correlate better with human evaluators as compared to the standard metrics. 

Generally, data augmentation in NLP is not entirely limited with producing very similar, and even context-relevant, responses as DA also has the potential to produce semantically-correct synthetic sequences that are contextually-distant to that of the original sequence. Hence, we introduce a third metric involving identifying the quantity of function words versus content words present in the response. Particularly, an increase in content words are theorized to be indicative of additions of detail in the generation. Function words traditionally refer to the articles, prepositions, pronouns, conjunction and other parts of speech with specific syntactic usage in constructing sentences, whereas content words are the nouns, verbs, adjectives, and some adverbs \cite{MILLER1958370}. Analyzing large amounts of text in the same study, the average character length of a function word was identified to be at 3.13 characters, while content words had an average of 6.47 characters. With this in mind, tokens with 1-3 characters were categorized as function words and tokens with 4-15 characters were categorized as content words. Tokens that are solely made up of punctuation were not included during processing. The slight adjustment of character counts was done to account for the lengthier Tagalog words observed in the dataset.

\section{Data Preparation}

PEx data was used in both finetuning a Tagalog RoBERTa model for DA and the DialoGPT model. We used a 97\%-3\% split for DialoGPT and RoBERTa respectively. To produce the splits, we load the full dataset into Pandas and set the random seed to 42. 

Conversation-like exchanges were derived by extracting depth-first chains of the structure topic-reply-reply in the recursive structure, creating conversations. 5\% from the DialoGPT split was taken to serve as the RoBERTa evalaution set. The remaining was used in the data augmentation to produce the synthetic data, which was then also used in conjunction with the original set when training the DialoGPT model. This set was split into 80\%-20\% for the training set and test set. The training set goes through the data augmentation process for generation of synthetic data while the test set remains untouched.

\section{Data Augmentation}
\label{sec:dataaug}

We fine-tuned a RoBERTa model in order to capture the nuances of casual written Filipino present in the dataset. We used Tagalog RoBERTa \cite{cruz2021improving} then fine-tuned it with 3\% of the PEx dataset split into an 80\%-20\% test and validate set, with a split random seed of 42. We fine-tuned RoBERTa for 3 epochs, using a batch size of 8, and a learning rate of 5e-5. 

To increase the size of the existing corpora, the flattened conversations from preprocessing are separated into independent utterances. This will allow our Tagalog RoBERTa model to perform mask-fill on a per-response basis over a per-conversation basis. 

A percentage is indicated for augmentation in order to limit the amount of original tokens that are replaced with RoBERTa-predicted tokens. To ensure that the amount of token replacements scales according to the sequence length of each utterance, the specified percentage is multiplied with each sequence length where a ceiling function is used to return the smallest succeeding integer (i.e. 15\% of 10 tokens equate to 2 tokens to be replaced).

This predetermined amount was used to randomly select $n$ amount of indices corresponding to the list of tokens. In a sequential manner, the token in each selected index in the original sequence was replaced with a \texttt{<mask>} token. The altered sequence was then fed to the fine-tuned RoBERTa model in order to predict a replacement in place of the mask token, choosing the highest scoring prediction on every repetition. The process of masking and filling was iteratively performed until all $n$ amount of indices are consumed. The RoBERTa-predicted tokens were used to replace their corresponding tokens in the original sequence, hereby forming the synthetic utterance.

Each synthetic utterance was then merged along with the other generated utterances to form a synthetic conversation, effectively doubling the training corpus. Both the synthetic conversations were compiled with the original conversations to form a 50/50 dataset for finetuning DialoGPT.

\section{Response Generation Model}
\import{./tables/}{table5-6.tex}

\import{./tables/}{table5-7.tex}

Our model of choice for all our experiments is DialoGPT-medium (345M). We fine-tune DialoGPT for 5 epochs, using a batch size of 4, and a learning rate of 5e-5. 

Finetuning DialoGPT requires a dataset where each conversation has an equal length of exchanges. This essentially means that it needs to learn that the latest, or $n$th turn in that exchange takes into account the previous $n-1$ turns for its $n$th reply. Due to the varying length of exchanges present in the PEx data, we designated a specific length of 4 turns for all exchanges. Conversations with less than 4 exchanges were dropped. For conversations with more than 4 exchanges, contiguous sequences of 4 were derived from the beginning of the conversation until the last response. This means that a conversation with $x$ exchanges can generate $x-n+1$ contiguous sequences of $n$, e.g. a conversation with exchanges $e1, e2, e3, e4, e5$, we can generate the conversations (1) $e1, e2, e3, e4$, and (2) $e2, e3, e4, e5$. If an exchange happens to get filtered out in preprocessing, the whole sequence is dropped as there will be a gap in context.

Our fine-tuned model is capable of generating responses based on the previous 3 responses. Therefore, the model is presented with a concatenated sequence of these responses delimited with an end-of-sequence token. 

The decoding method used for both manual and automatic testing is beam search of \textit{width} 5. A \textit{trigram} repetition penalty was also introduced due to some instances where beam search would produce a repetitive token generation causing the max sequence length to be fully consumed. We saw the need for applying such penalty as it directly affected hypothesis generation.

\section{Finetuning Results}

For data augmentation, we fine-tuned the existing Tagalog RoBERTa model with 3\% of the whole dataset, equating to 178K individual utterances, for 3 epochs in approximately 40 minutes. The model resulted in a perplexity of 17.7287.

For DialoGPT, we fine-tuned a baseline model and five augmented models with varying percentages of token replacement. The baseline model was fine-tuned with all 53K untouched conversations from the corpus and zero synthetic data for approximately 10 hours. The augmented models were fine-tuned with a merged corpus of original data (53K) and synthetic data (53K) for approximately 18 hours each. The varying percentages of RoBERTa-generated tokens present in the synthetic dataset constitutes the only independent variable for the augmented models. We maintain identical training configurations for all the experiments. The specific finetuning scores are shown in Table \ref{tab:iter1modelsummary}.

As part our ablations below, we fine-tuned another six models with decreased training data sizes following the same training configurations. Finetuning results are displayed in Table \ref{tab:iter2modelsummary}.

\section{Ablation Results and Discussion}
In this section, we perform ablations to understand the effects of the data augmentation percentages with regards to model performance. We also simulated lower training data sizes to assess effectiveness of the methodology. In addition, sample conversations within and outside the domain of the corpus are also presented in this section. 

\subsection{Effect of Token Replacement Percentage}

An ablation was performed to determine the effect of varying token replacement percentages on the final performance of the model. The evaluation scores can be seen in Table \ref{tab:iter1-eval}

As shown in Table \ref{tab:iter1-eval}, replacing 10\% of the original tokens exhibits the highest improvement in model performance both in Perplexity and BERTScore as compared to the baseline model. We observe a degradation in both metrics after the 10\% token replacement when increasing the replacement percentage further. We theorized that the presence of more synthetic tokens introduces less contextually-similar responses regardless if the synthetic tokens are semantically valid. We also hypothesize that there is a higher possibility of semantically-invalid tokens present in higher replacement percentages as the masking and filling process occurs independently as mentioned in Section \ref{sec:dataaug}. 

We also sought to determine the effect of the token replacement percentages outside context similarity and probability. We observe a significant difference in the amount of content words and functions words for all augmented models. This provides evidence that synthetic data introduces lengthier responses in general. More importantly, the increase in content words reinforces our observation that the augmented models includes more substance with regards to detail in the generated response.  

\import{./tables/}{table5-1.tex}
\import{./tables/}{table5-2.tex}

\subsection{Effect of Training Data Size}

Using 10\% token replacement, we also trained six models across varying data sizes to determine the effect of the methodology in other scales of low data sizes.
The model with a base training size of 1K samples improved the most in terms of perplexity achieving a decrease of 1.103. The full training size also improved by a significant margin of 10.75\% change, albeit only a 0.4435 decrease in perplexity. Scores for the 53K models were taken from the experiments on token replacement percentage.

The models also showed relative minor improvements in BERTScore as seen in Table \ref{tab:iter2autoeval}. The model with a base training size of 1K samples improved in its BERTScore as well by 1.59\% or a 0.0054 increase in F1 after being trained with synthetic data. The full training size improved the most, with a 3.26\% increase in F1, while the models trained with 10K and 25K base training sizes improved relatively insignificant. 

We theorized that the smaller training samples tend to be more sensitive when using a model such as DialoGPT medium which requires more data on average than 1K samples. We speculate that the dip in scores for 10K and 25K base training sizes and the sudden increase in scores again for the full 53K training size could be attributed to the normal expected training sizes of the model. Consistent with the previous ablation, content word and function word usage also increased with all augmented models as seen in Table \ref{tab:iter2autoeval}.

\section{Conclusion}
We constructed the first Filipino conversational dataset called PEx Conversations. We leverage this dataset by proposing a RoBERTa-based data augmentation methodology via a fine-tuned Tagalog RoBERTa model to perform context-aided token replacement. The collected data and the synthetic data were both used to fine-tune DialoGPT to produce the first Filipino conversational response generator capable of generating a response related to the previous 3 responses.

We were able to show that introducing synthetic data does improve model performance by a comparable margin. The synthetic data was able to enhance our best model in BERTScore F1 Accuracy by 12.2\% and perplexity by 10.7\%.  We also found that introducing synthetic data also increases the usage of content words in a response, with our highest scoring augmented model achieving a 11.7\% increase. 

While the ablations show that replacing 10\% tokens in an utterance improved the performance quantitatively, the qualitative differences between having more tokens replaced were negligible as observed. Additionally, we also experimented with applying our technique with simulated lower training data sizes. Our findings show that the technique still relatively improves performance using our 1K, 10K, and 25K training size samples. 

To validate the effectiveness of our DA, we suggest looking into applying the method on other Filipino conversational datasets, or even recreate the technique with other low-resource languages if possible. Chat data, in contrast to transformed forum data, can also be explored when attempting to produce more conversation-like responses.

For DA, our methodology involves independent replacement followed by a merge step when producing synthetic utterances. Hence, we recommend exploring into a cascading approach where replacement occurs in succession where the newly-produced sequence is used as the candidate for the next token replacement.

With regards to the model, it can also be beneficial to look into finetuning the large DialoGPT model (762M) for corpora of a larger-scale. In order to reinforce the findings, we also encourage future work to perform human evaluations to understand the effect of the DA qualitatively, and potentially with the support of expert linguists should the resources allow. 

Lastly, we recommend exploring real-world applications of these conversational agents. This includes agents that may have different personas or conversing personalities, to suit specific purposes such as being a companion, an assistant or an expert.

\section{Bibliographical References}\label{reference}

\bibliographystyle{lrec2022-bib}
\bibliography{lrec2022-example}

\end{document}

%% file: tables/table5-6.tex
\begin{table*}[!ht]
    \centering
    \begin{tabular}{ |c|c|c|c|c| }
        \hline
        \textbf{\%} & \textbf{Data Size} & \textbf{Avg Loss} & \textbf{Training Time in Hrs} & \textbf{Prplx}  \\
        \hline
        0\% & 52,640 & 1.5436 & 9.75 & 4.1206\\ \hline
        10\% & 105,280 & 1.5306 & 18.2 & 3.6771\\ \hline
        15\% & 105,280 & 1.6059 & 18 & 3.7867\\ \hline
        20\% & 105,280 & 1.6646 & 18.2 & 3.8593\\ \hline
        25\% & 105,280 & 1.7156 & 18.8 & 3.9142\\ \hline
        30\% & 105,280 & 1.7610 & 17.8 & 3.9574\\ 
        \hline
    \end{tabular}
    \caption{\label{tab:iter1modelsummary} Results of fine-tuning models with varying percentages of token replacement}
\end{table*}

%% file: tables/table5-7.tex
\begin{table*}[!h]
    \centering
    \begin{tabular}{ |c|c|c|c|c| }
        \hline
        Method & Data Size & Avg Loss & Training Time in Hrs & Prplx  \\
        \hline
        1K Base & 1,000 & 2.4788 & 0.17 & 9.5821\\ \hline
        1K Aug & 2,000 & 2.3005 & 0.35 & 8.4791\\ \hline \hline
        10K Base & 10,000 & 1.8721 & 1.6 & 6.1937\\ \hline
        10K Aug & 20,000 & 1.8311 & 3.4 & 5.755\\ \hline \hline
        25K Base & 25,000 & 1.7009 & 4.3 & 4.9905\\ \hline
        25K Aug & 50,000 & 1.6780 & 9.16 & 4.5684\\ \hline
    \end{tabular}
    \caption{\label{tab:iter2modelsummary} Results of fine-tuning models in varying training data sizes}
\end{table*}

%% file: tables/table5-1.tex
\begin{table}[!h]
\centering
\begin{tabular}{|l|llll|}
    \hline \textbf{\%} & \textbf{Prplx} & \textbf{F1} & \textbf{Func.} & \textbf{Cont.} \\ \hline
    
    0\% & 4.1206 & 0.3379 & 82,956 & 100,940  \\ \hline
    
    10\% & 3.6771 & 0.3794 & 88,944 & 112,776 \\ \hline
    
    15\% & 3.7867 & 0.3762 & 88,175 &  116,501 \\ \hline

    20\% & 3.8593 & 0.3740 & 87,910 & 114,018  \\ \hline

    25\% & 3.9142 & 0.3719 & 88,467 & 113,473 \\ \hline
    
    30\% & 3.9574 & 0.3704 & 86,856 & 110,135 \\ \hline

\end{tabular}
\caption{\label{tab:iter1-eval}Perplexity, F1, and Function Word and Content Word count for the models with different token replacement percentages}
\vspace{-4mm}
\end{table}

%% file: tables/table5-2.tex
\begin{table}[!h]
\centering
\begin{tabular}{|l|lllp{1.1cm}|}
    \hline \textbf{Setup} & \textbf{Prplx} & \textbf{F1} & \textbf{Func.} & \textbf{Cont.} \\ \hline
    
    1K & 9.5821 & 0.3387 & 57,327 & 58,872  \\ 
    
    1K+Aug & 8.4791 & 0.3441 & 60,774 & 64,368 \\ \hline
    
    10K & 6.1937 & 0.3520 & 66,371 &  71,706 \\ 

    10K+Aug & 5.755 & 0.3539 & 70,078 & 80,662  \\ \hline

    25K & 4.1206 & 0.3584 & 76,665 & 87,873  \\ 
    
    25K+Aug & 3.6771 & 0.3617 & 82,240 & 104,211  \\ \hline

\end{tabular}
\caption{\label{tab:iter2autoeval}Perpelixty, F1, and Function Word and Content Word count for varying data sizes}
\end{table}

%% file: lrec2022-example.bbl
\begin{thebibliography}{}

\bibitem[\protect\citename{Adiwardana \bgroup et al.\egroup
  }2020]{adiwardana2020humanlike}
Adiwardana, D., Luong, M.-T., So, D.~R., Hall, J., Fiedel, N., Thoppilan, R.,
  Yang, Z., Kulshreshtha, A., Nemade, G., Lu, Y., and Le, Q.~V.
\newblock (2020).
\newblock Towards a human-like open-domain chatbot.

\bibitem[\protect\citename{Baheti \bgroup et al.\egroup
  }2020]{baheti-etal-2020-fluent}
Baheti, A., Ritter, A., and Small, K.
\newblock (2020).
\newblock Fluent response generation for conversational question answering.
\newblock In {\em Proceedings of the 58th Annual Meeting of the Association for
  Computational Linguistics}, pages 191--207, Online, July. Association for
  Computational Linguistics.

\bibitem[\protect\citename{Cruz and Cheng}2021]{cruz2021improving}
Cruz, J. C.~B. and Cheng, C.
\newblock (2021).
\newblock Improving large-scale language models and resources for filipino.
\newblock {\em arXiv preprint arXiv:2111.06053}.

\bibitem[\protect\citename{Cruz \bgroup et al.\egroup
  }2021]{cruz2021exploiting}
Cruz, J. C.~B., Resabal, J.~K., Lin, J., Velasco, D.~J., and Cheng, C.
\newblock (2021).
\newblock Exploiting news article structure for automatic corpus generation of
  entailment datasets.
\newblock In {\em Pacific Rim International Conference on Artificial
  Intelligence}, pages 86--99. Springer.

\bibitem[\protect\citename{Dai and Adel}2020]{Dai2020}
Dai, X. and Adel, H.
\newblock (2020).
\newblock An analysis of simple data augmentation for named entity recognition.

\bibitem[\protect\citename{Devlin \bgroup et al.\egroup
  }2019]{devlin-etal-2019-bert}
Devlin, J., Chang, M.-W., Lee, K., and Toutanova, K.
\newblock (2019).
\newblock {BERT}: Pre-training of deep bidirectional transformers for language
  understanding.
\newblock In {\em Proceedings of the 2019 Conference of the North {A}merican
  Chapter of the Association for Computational Linguistics: Human Language
  Technologies, Volume 1 (Long and Short Papers)}, pages 4171--4186,
  Minneapolis, Minnesota, June. Association for Computational Linguistics.

\bibitem[\protect\citename{Goyal \bgroup et al.\egroup
  }2018]{goyal-etal-2018-fast}
Goyal, A.~K., Metallinou, A., and Matsoukas, S.
\newblock (2018).
\newblock Fast and scalable expansion of natural language understanding
  functionality for intelligent agents.
\newblock In {\em Proceedings of the 2018 Conference of the North {A}merican
  Chapter of the Association for Computational Linguistics: Human Language
  Technologies, Volume 3 (Industry Papers)}, pages 145--152, New Orleans -
  Louisiana, June. Association for Computational Linguistics.

\bibitem[\protect\citename{Grundkiewicz \bgroup et al.\egroup
  }2019]{grundkiewicz-etal-2019-neural}
Grundkiewicz, R., Junczys-Dowmunt, M., and Heafield, K.
\newblock (2019).
\newblock Neural grammatical error correction systems with unsupervised
  pre-training on synthetic data.
\newblock In {\em Proceedings of the Fourteenth Workshop on Innovative Use of
  NLP for Building Educational Applications}, pages 252--263, Florence, Italy,
  August. Association for Computational Linguistics.

\bibitem[\protect\citename{Kobayashi}2018]{kobayashi-2018-contextual}
Kobayashi, S.
\newblock (2018).
\newblock Contextual augmentation: Data augmentation by words with paradigmatic
  relations.
\newblock In {\em Proceedings of the 2018 Conference of the North {A}merican
  Chapter of the Association for Computational Linguistics: Human Language
  Technologies, Volume 2 (Short Papers)}, pages 452--457, New Orleans,
  Louisiana, June. Association for Computational Linguistics.

\bibitem[\protect\citename{Liu \bgroup et al.\egroup }2019]{liu2019roberta}
Liu, Y., Ott, M., Goyal, N., Du, J., Joshi, M., Chen, D., Levy, O., Lewis, M.,
  Zettlemoyer, L., and Stoyanov, V.
\newblock (2019).
\newblock Roberta: A robustly optimized bert pretraining approach.

\bibitem[\protect\citename{Miller \bgroup et al.\egroup }1958]{MILLER1958370}
Miller, G., Newman, E., and Friedman, E.
\newblock (1958).
\newblock Length-frequency statistics for written english.
\newblock {\em Information and Control}, 1(4):370--389.

\bibitem[\protect\citename{Puri \bgroup et al.\egroup
  }2020]{puri-2020-TrainingQA}
Puri, R., Spring, R., Patwary, M.~A., Shoeybi, M., and Catanzaro, B.
\newblock (2020).
\newblock Training question answering models from synthetic data.
\newblock {\em ArXiv}, abs/2002.09599.

\bibitem[\protect\citename{See \bgroup et al.\egroup }2017]{see-etal-2017-get}
See, A., Liu, P.~J., and Manning, C.~D.
\newblock (2017).
\newblock Get to the point: Summarization with pointer-generator networks.
\newblock In {\em Proceedings of the 55th Annual Meeting of the Association for
  Computational Linguistics (Volume 1: Long Papers)}, pages 1073--1083,
  Vancouver, Canada, July. Association for Computational Linguistics.

\bibitem[\protect\citename{Vania \bgroup et al.\egroup
  }2019]{vania-etal-2019-systematic}
Vania, C., Kementchedjhieva, Y., S{\o}gaard, A., and Lopez, A.
\newblock (2019).
\newblock A systematic comparison of methods for low-resource dependency
  parsing on genuinely low-resource languages.
\newblock In {\em Proceedings of the 2019 Conference on Empirical Methods in
  Natural Language Processing and the 9th International Joint Conference on
  Natural Language Processing (EMNLP-IJCNLP)}, pages 1105--1116, Hong Kong,
  China, November. Association for Computational Linguistics.

\bibitem[\protect\citename{Wei and Zou}2019]{Wei-etal-2019-EDA}
Wei, J.~W. and Zou, K.
\newblock (2019).
\newblock {EDA:} easy data augmentation techniques for boosting performance on
  text classification tasks.
\newblock {\em CoRR}, abs/1901.11196.

\bibitem[\protect\citename{Wu \bgroup et al.\egroup }2018]{wu-etal-2018-bert}
Wu, X., Lv, S., Zang, L., Han, J., and Hu, S.
\newblock (2018).
\newblock Conditional {BERT} contextual augmentation.
\newblock {\em CoRR}, abs/1812.06705.

\bibitem[\protect\citename{Xu \bgroup et al.\egroup }2017]{xu2017new}
Xu, A., Liu, Z., Guo, Y., Sinha, V., and Akkiraju, R.
\newblock (2017).
\newblock A new chatbot for customer service on social media.
\newblock In {\em Proceedings of the 2017 CHI Conference on Human Factors in
  Computing Systems}, pages 3506--3510.

\bibitem[\protect\citename{Zhang \bgroup et al.\egroup
  }2019]{zhang2019dialogpt}
Zhang, Y., Sun, S., Galley, M., Chen, Y.-C., Brockett, C., Gao, X., Gao, J.,
  Liu, J., and Dolan, B.
\newblock (2019).
\newblock Dialogpt: Large-scale generative pre-training for conversational
  response generation.
\newblock {\em arXiv preprint arXiv:1911.00536}.

\bibitem[\protect\citename{Zhang \bgroup et al.\egroup
  }2020]{zhang2020bertscore}
Zhang, T., Kishore, V., Wu, F., Weinberger, K.~Q., and Artzi, Y.
\newblock (2020).
\newblock Bertscore: Evaluating text generation with bert.

\end{thebibliography}
